\documentclass[runningheads]{llncs}
\usepackage{graphicx}
\usepackage{amsmath,amssymb} 
\usepackage{color}
\usepackage[width=122mm,left=12mm,paperwidth=146mm,height=193mm,top=12mm,paperheight=217mm]{geometry}
\usepackage{multirow}
\begin{document}
\pagestyle{headings}
\mainmatter

\title{Learnable Histogram: Statistical Context Features for Deep Neural Networks} 

\titlerunning{Learnable Histogram: Statistical Context Features for Deep Neural Networks}

\authorrunning{Zhe Wang, Hongsheng Li, Wanli Ouyang, Xiaogang Wang}


\author{Zhe Wang, Hongsheng Li\thanks{Corresponding authors}, Wanli Ouyang, Xiaogang Wang$^{\star}$}
\institute{$^\dagger$Department of Electronic Engineering\\The Chinese University of Hong Kong, Hong Kong \\$^\ddagger$SenseNets Technology Ltd. \\ \texttt{\{zyzhao, hsli, xgwang\}@ee.cuhk.edu.hk}, \texttt{zhaorui@sensenets.com}} 
\institute{ Dept. of  Electronic Engineering, The Chinese University of Hong Kong\\
        \email{ \{zwang, hsli, wlouyang, xgwang\}@ee.cuhk.edu.hk}
}


\maketitle

\begin{abstract}
Statistical features, such as histogram, Bag-of-Words (BoW) and Fisher Vector, were commonly used with hand-crafted features in conventional classification methods, but attract less attention since the popularity of deep learning methods. In this paper, we propose a learnable histogram layer, which learns histogram features within deep neural networks in end-to-end training. Such a layer is able to back-propagate (BP) errors, learn optimal bin centers and bin widths, and be jointly optimized with other layers in deep networks during training. Two vision problems, semantic segmentation and object detection, are explored by integrating the learnable histogram layer into deep networks, which show that the proposed layer could be well generalized to different applications. In-depth investigations are conducted to provide insights on the newly introduced layer.
\end{abstract}

\section{Introduction}

\label{sec:intro}
\begin{figure}[t]
\centering
  \includegraphics[height=4.5cm]{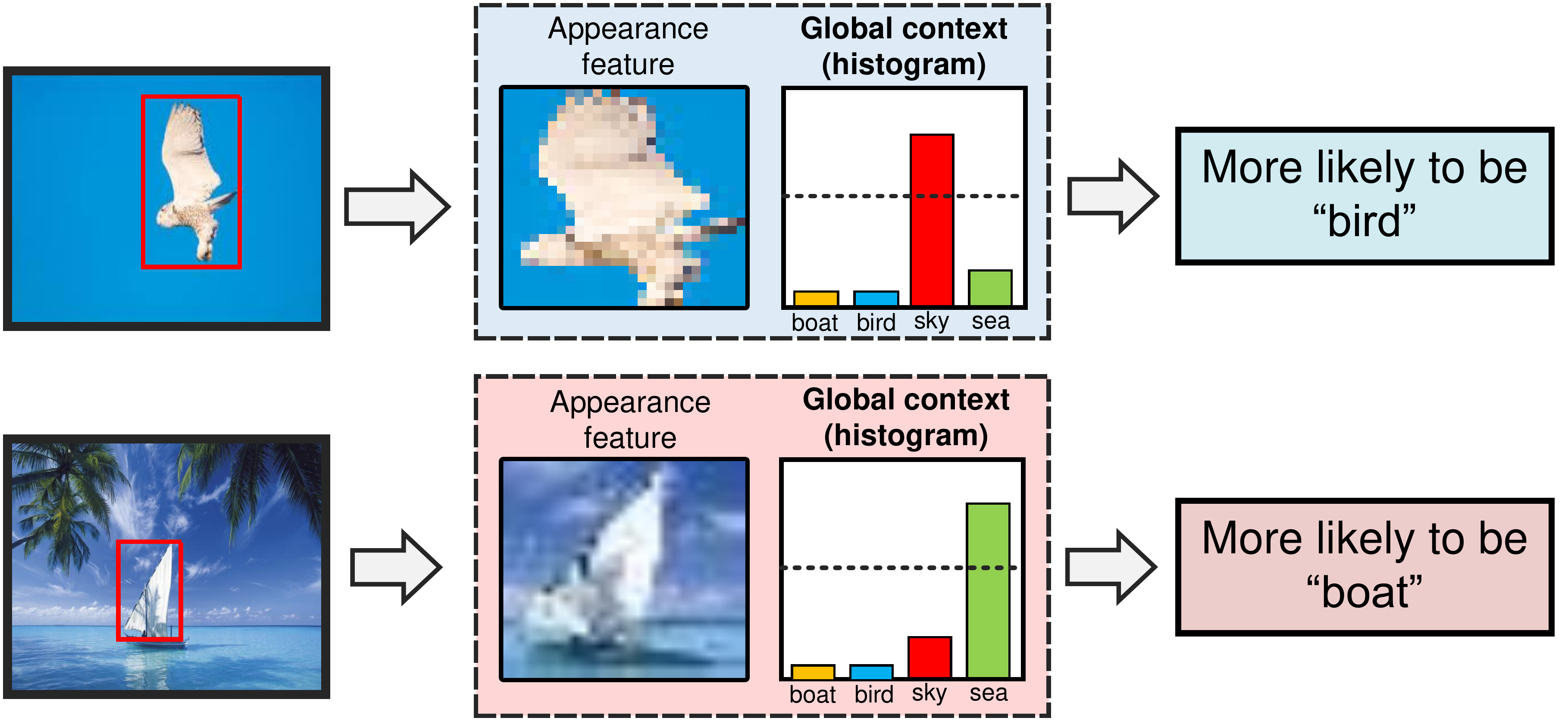}\\
   \caption{A toy example showing that the global context (histogram) of a whole image is helpful for classifying image patches. The image patch is more likely to be a ``bird'' if the histogram has higher bin counts on the class ``sky'', or a ``boat'' if the histogram has higher bin counts on the class ``sea''. }\label{fig:intro}
\end{figure} 
Context features play a crucial role in many vision classification problems, such as semantic segmentation  \cite{yang2014context,gould2009decomposing,barinova2010geometric,ladicky2010graph,shotton2006textonboost,yao2012describing}, object detection \cite{szegedy2014scalable,ouyang2014deepid} and pose estimation \cite{fan2015combining,carreira2015human}. As illustrated by the toy example in Fig. \ref{fig:intro}, when performing classification on the blurry white objects with similar appearance, if the semantic histogram from the whole image has a higher bin on the class ``sea'', then the object is more likely to be classified as a ``boat''; if the histogram has a higher bin on the class ``sky'', then it is more likely to be classified as a ``bird''.
The semantic context thus acts as an important indicator for this classification task. 

Context features could be mainly categorized into statistical and non-statistical ones depending on whether they abandon the spatial orders of the context information.
On the one hand, for most deep learning methods that gain increasing attention in recent years, non-statistical context features dominate. Some examples include \cite{rirshick2014rich} for object detection and  \cite{farabet2013learning} for semantic segmentation.

On the other hand, statistical context features were mostly used in conventional classification methods with hand-crafted features. Commonly used statistical features include histogram, Bag-of-Words (BoW) \cite{lazebnik2006beyond}, Fisher vector \cite{perronnin2010improving}, Second-order pooling \cite{carreira2012semantic}, etc. Such global context features performed successfully with hand-crafted low-level features at their times. However, they were much less studied since the popularity of deep learning. 
There are a limited number of deep learning methods that tried to incorporate statistical features into deep neural networks. Such examples include the deep Fisher network \cite{simonyan2013deep} that incorporate Fisher vector and orderless pooling \cite{gong2014multi} that combines with Vector of Locally Aggregated Descriptors (VLAD). Both methods aim to improve the image classification performance. However, when calculating the statistical features, both methods fix the network parameters and simply treat features by deep networks as off-the-shelf features. In such a way, the deep networks and the statistical operations are not jointly optimized, which is one of the key factors for the success of deep networks.
In this work, we introduce a learnable histogram layer for deep neural networks. 

Unlike existing deep learning methods that treat statistical operations as a separate module, our proposed histogram layer is able to back-propagate (BP) errors and learn optimal bin centers and bin width during training. Such properties make it possible to be integrated into neural networks and end-to-end trained. In this way, the appearance and statistical features in a neural network could effectively adapt each other and thus lead to better classification accuracy.

The proposed learnable histogram layer could be used for various applications.
We propose the HistNet-SS network for semantic segmentation and the HistNet-OD network for object detection. Both networks are built based on state-of-the-art deep neural networks with the learnable histogram layer. Jointly training the HistNets in an end-to-end manner helps convolution layers learn more discriminative feature representations and boosts the final accuracy.
Thus our contributions of this paper can be summarized as three-fold:
\begin{itemize}
\item We propose the learnable histogram layer for deep neural networks, which is able to BP errors, calculate histogram features, and learn optimal bin centers and widths. To the best of our knowledge, such learnable histogram features are introduced to deep learning for the first time.
\item We conduct thorough investigations on the proposed statistical feature with comparison to the non-statistical counterparts. We show that statistical features help achieve better accuracy with fewer parameters in certain cases.
\item We show that the proposed learnable histogram layer is easy to generalize and could be utilized for different applications. We proposed two HistNet models for solving semantic segmentation and object detection problems. State-of-the-art performance is achieved for both applications.
\end{itemize}

\section{Related work}

\subsubsection{Semantic segmentation by deep learning.} 
The goal of semantic segmentation is to densely classify pixels in a given image into one of the predefined categories. 
Recently, deep learning based methods have dramatically improved the performance of semantic segmentation. 
Farabet et al. \cite{farabet2013learning} proposed a multi-scale convolution neural network for semantic segmentation. Their model takes a large image patch as input to cover more context around the center pixel, and applies post-processing techniques such as superpixel voting and Conditional Random Field  (CRF) to improve the consistency of the labeling map. 
Pinheiro et al. \cite{pinheiro2013recurrent} used a Recurrent Neural Network (RNN) to recursively correct its own mistakes by taking the raw image together with the predictions of the previous stage as input. 
Long et al. \cite{long2014fcn} proposed the Fully Convolution Network (FCN) which takes the whole image as input and is trained in an end-to-end manner. Following \cite{long2014fcn}, many works have been proposed to incorporate more semantic context information into deep learning model. Chen et al. \cite{chen2014semantic} combined the output of the FCN with a fully connected CRF. However, the two components are treated as two separate parts and greedily trained.  Zheng et al. \cite{zheng2015conditional} showed that the mean-field algorithm for solving a fully connected CRF is equivalent to a RNN, which can be jointly trained with the FCN in an end-to-end manner. 
Liu et al. \cite{liu2015semantic} designed layers to model pairwise terms in a MRF, which approximate the mean-field by only one iteration, and thus makes inference much faster. 
Our work differ with these methods in the way that we model context as statistical features while they model context with graphical models.  These methods and our proposed method are complementary, and can be utilized in a unified framework to further improve the performance.

\subsubsection{Object detection by deep learning.}
The object detection aims at locating the objects of predefined categories in a given image. 
RCNN \cite{rirshick2014rich} is a famous pipeline based on CNN. It first pre-trains the CNN on the image classification task, and then uses the learned weights as the initial point for training the detection task with region proposals. Many works have been proposed to improve RCNN. The faster-RCNN \cite{renNIPS15fasterrcnn} simultaneously predicts the region proposals  and outputs the detection scores in a given image, while the two parts share the same convolution layers. Although their model takes the whole image as input, the global context information is ignored. Ouyang et al. \cite{ouyang2014deepid} used the image classification scores from another CNN as the semantic context information to refine the detection scores of the bounding boxes produced by the RCNN pipeline. Szegedy et al. \cite{szegedy2014scalable} concatenated the topmost features of image classification to those of all the detection bounding boxes. However, both methods require extra training data and labels on the image classification task.  In comparison, our work calculates the likelihood histogram of the base model's own prediction as global context, which does not require any extra annotation.

\subsubsection{Statistical features in deep learning.}
Some other works have been proposed to incorporate statistical models into a deep learning framework. Simonyan et al. \cite{simonyan2013deep} proposed a Fisher Vector Layer, which is the generalization of a standard Fisher Vector, and a Fisher Vector network, which consists of a stack of Fisher Vector Layers. However, they still use conventional hand-crafted features as input of the network.  Gong \cite{gong2014multi}  presented a multi-scale orderless pooling scheme to extract global context features for image classification and image retrieval tasks.  They adopted the Vector of Local Aggregated Descriptors (VLAD) for encoding activations from a convolution neural network. However, unlike our learnable histogram layer layer, their model is not differentiable thus unable to BP errors.

\subsubsection{Differentiable histograms.}\label{subsubscn:differentiable}
Chiu et al. \cite{chiu2015see} exploited the pipeline of Histogram of Oriented Gradient (HOG) descriptor and showed it is piecewise differentiable. The key differences between our proposed layers and the differentiable HOG are three-fold. 1) Our learnable histogram layer does not only BP errors but also learns optimal bin centers and bin widths during training, while the differentiable HOG has fixed bin centers and widths. 2) We for the first time introduce the learnable histogram layer into deep neural networks for end-to-end training. All the learnable layers in a neural network could effectively adapt each other for learning better feature representations. 3) We also show that such a learnable histogram layer could be formulated by a stack of existing CNN layers and thus significantly lowers the implementation difficulty.

\section{Methods}
\label{sec:methods}
\subsection{The overall frameworks}
Global semantic context has been shown of great effectiveness in various classification problems including semantic segmentation \cite{yang2014context,gould2009decomposing,barinova2010geometric,ladicky2010graph,shotton2006textonboost,yao2012describing}, object detection \cite{szegedy2014scalable,ouyang2014deepid} and pose estimation \cite{fan2015combining,carreira2015human}. Histogram is one of the most-commonly-used conventional statistical features for describing context. However, such statistical features are little investigated  by existing deep learning methods.

\begin{figure}[t]
\centering
  \includegraphics[height=3.4cm]{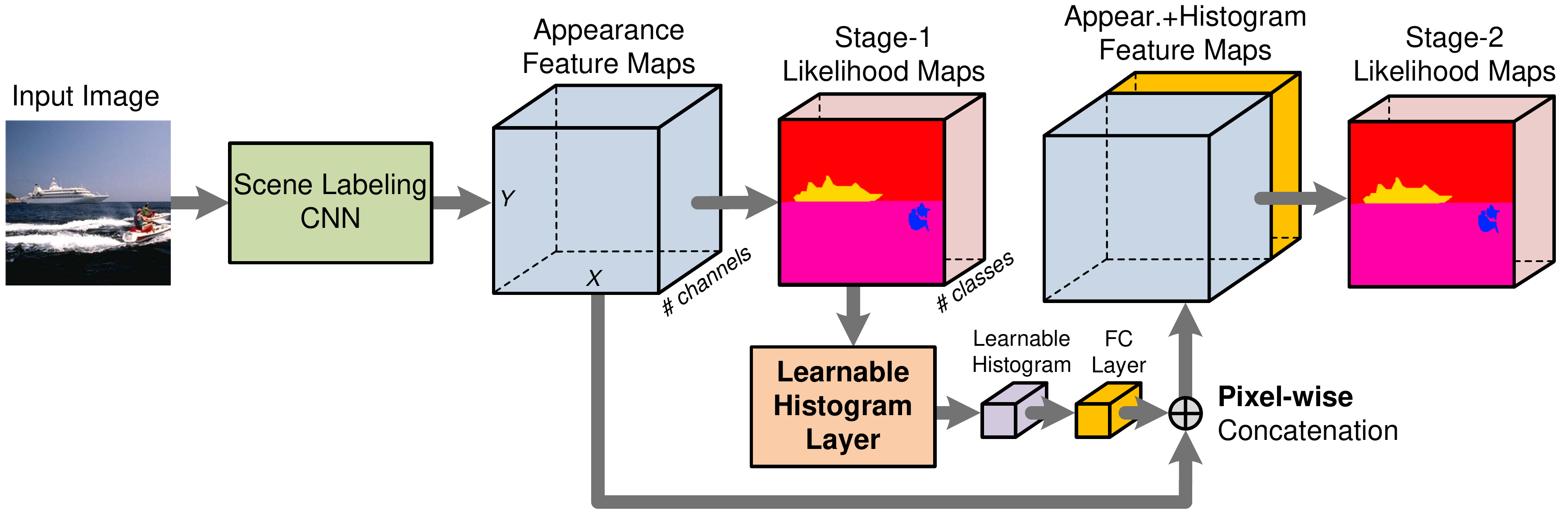}\\
  (a) HistNet-SS: the proposed network for semantic segmentation.\\
   \includegraphics[height=2.5cm]{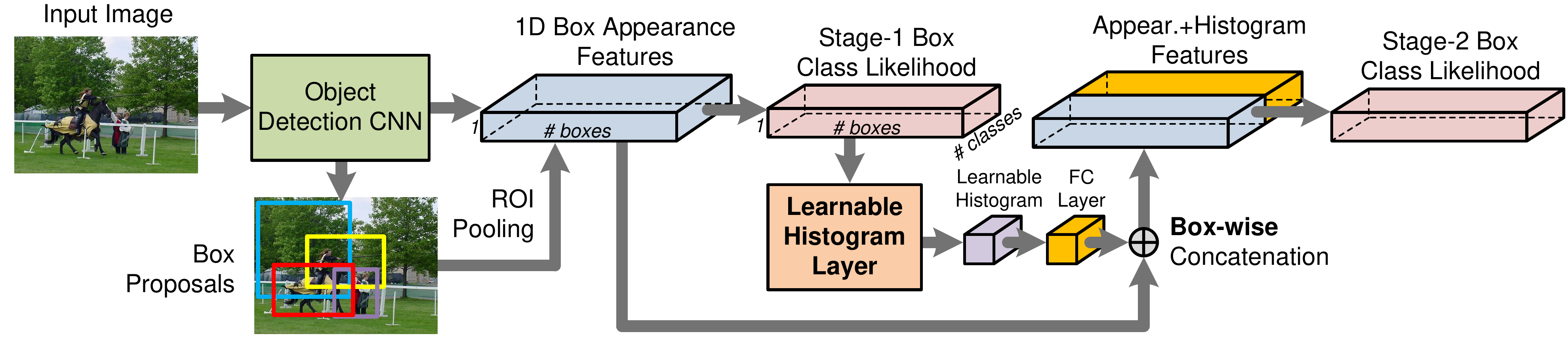}\\(b) HistNet-OD: the proposed network for object detection.
   \caption{The pipelines of the proposed HistNet-SS for semantic segmentation and HistNet-OD for object detection. A learnable histogram layer is included to incorperate the global semantic context inforamtion. }\label{fig:framework}
\end{figure} 

We propose two deep neural networks, the HistNet-SS for semantic segmentation and the HistNet-OD for object detection. Both include a learnable histogram layer that calculates histogram features for a likelihood map or vector. The learnable histogram layer can BP errors to bottom layers, and automatically learns the optimal bin centers and bin widths. Such properties make it possible to be trained  in a deep neural network in an end-to-end manner.

The semantic segmentation task requires labeling each pixel of an input image with a given class. Our HistNet-SS is based on the FCN-VGG network \cite{long2014fcn}, which takes a whole image as input and outputs all pixels' class likelihood simultaneously. As shown in Fig. \ref{fig:framework}(a), our proposed HistNet-SS model adds a learnable histogram layer following the initial class likelihood map (stage-1 likelihood map) by the FCN-VGG to obtain the histogram features of the likelihood map of the whole image. Such histogram features are then forwarded through a
Fully Connected (FC) layer and pixel-wisely concatenated with the topmost feature maps of the FCN-VGG. A new $1 \times 1$ convolution layer is added as the stage-2 classifier to generate the stage-2 likelihood map for each pixel of the input image. In this way, the global semantic context could provide vital information when classifying each pixel. For instance, for an image in the SIFTFlow dataset \cite{liu2008sift}, if the histogram shows that the ``sea''class has very large counts of high likelihoods, then the probability of classifying the pixels as ``street light'' should be diminished to some extent. The likelihood maps as output of stage-2 classifier can form a new histogram, which can be concatenated with the appearance features again to refine prediction in stage-3 and so on. The final class likelihood map is calculated as the average of the likelihood maps at all stages. During training, the supervision signals are applied to all the likelihood maps.

For the object detection task, each object of interest in an input image is required to be annotated by a bounding box with a confidence score. Our HistNet-OD is based on the faster-RCNN model \cite{renNIPS15fasterrcnn}, which includes a Region Proposal Network (RPN) and a fast-RCNN detector. For each input image, the RPN generates region proposals and the fast-RCNN detector extracts features for each region from the topmost feature map via ROI pooling and predicts the likelihoods of each box proposal belonging to pre-defined classes. Similar to our HistNet-SS model, our network feeds the initial box class likelihood (stage-1 box class likelihood) to our learnable histogram layer. The output histogram features encode statistics of the prediction class likelihood for the input image and then go through a FC layer. The resulting context features are then box-wisely concatenated with each box proposal's appearance features and classified by a fully connected layer to generate the stage-2 box class likelihood. The supervision signals are applied to all likelihood vectors, and the final likelihood are obtained by averaging those of the multiple stages.

\subsection{The learnable histogram layer}\label{subscn:adpt-hist}
\subsubsection{Conventional histograms.}

For the semantic segmentation or the object detection task, each sample (either a pixel or a box proposal) is labeled with $K$ scores by a neural network to denote its confidences on the $K$ pre-defined classes.  For calculating a conventional histogram on the samples' class scores, we divide each class score into $B$ bins, and the histogram is therefore of size $K\times B$. Each of the sample's $K$ class likelihoods casts a vote to its corresponding bin, and all bins' votes are then normalized to obtain the conventional histogram. The voting process for each bin of the conventional histogram could be treated as an indication function, which either votes $1$ (belonging to the bin) or $0$ (not belonging to the bin) for a specific sample. Those functions are not differentiable and cannot be utilized in a deep neural network for end-to-end training.

\begin{figure}[t]
\centering
\begin{tabular}{c@{\hspace{2mm}}c@{\hspace{1mm}}c@{\hspace{1mm}}c@{\hspace{1mm}}c}
   \includegraphics[height=2.2cm]{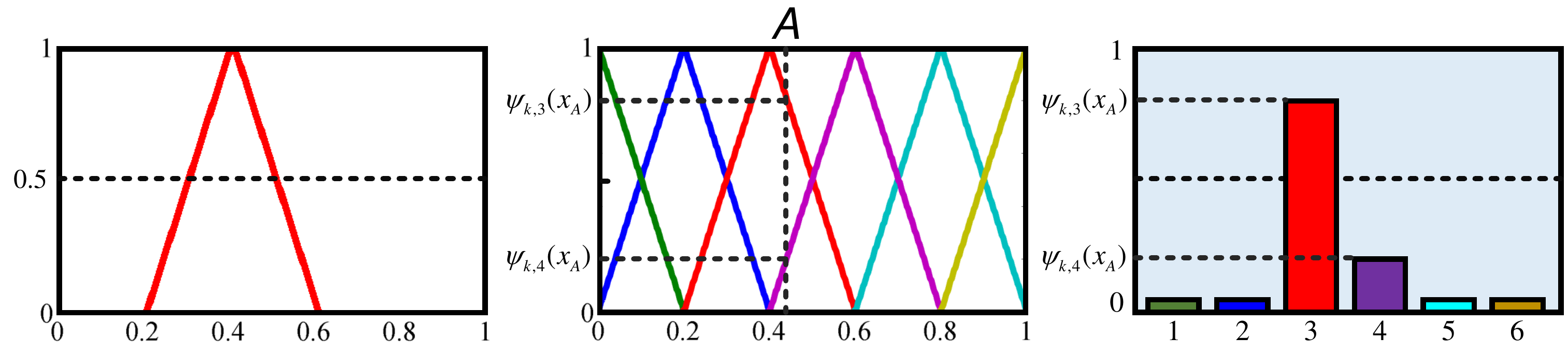}& \\
   (a) ~~~~~~~~~~~~~~~~~~~~~~~~~  (b)~~~~~~~~~~~~~~~~~~~~ ~~~~~ (c)\\
\end{tabular}
   \caption{(a) An example of the histogram basis function $\psi_{k,b}$. It corresponds to bin $3$ of the histogram functions in (b). (b) The histogram basis functions for class $k$ with $B=6$ bins. The sample $A$ with likelihood $x_A$ would vote for bin $3,4$ with non-negative weights $\psi_{k,3}(x_A)$ and $\psi_{k,4}(x_A)$. (c) The histogram of sample $A$'s class-$k$ likelihoods. }\label{fig:histogram}
\end{figure} 

\subsubsection{Learnable histograms.} Inspired by the differentiable Histogram of Oriented Gradient (HOG) in \cite{chiu2015see}, we design the learnable histogram layer for deep neural networks, which is piecewise differentiable and is able to BP errors. The differences between our work and \cite{chiu2015see} are summarized in Section \ref{subsubscn:differentiable}.

The $b$th bin of class $k$ in the learnable histogram is modeled by a piecewise linear basis function $\psi_{k,b}(x_k)$, (Fig. \ref{fig:histogram}. (a))
\begin{align}
\psi_{k,b}(x_k) =  \max\left\lbrace0,1-  \dfrac{1}{w_{k,b}} \times \vert x_k- \mu_{k,b}\vert  \right\rbrace, \label{fcn:hist}
\end{align}
where $\mu_{k,b}$ is its $b$th bin center for class $k$, $w_{k,b}$ is the bin width, and $\max\{\cdot, \cdot\}$ takes the maximum of the two values. Both the bin centers $\mu_{k,b}$ and bin widths $w_{k,b}$ could be learned during training. If a sample's class-$k$ score $x_k$ falls into the $b$th bin, i.e., the interval of $[\mu_b - w_{k,b}, \mu_b + w_{k,b}]$, this sample votes for the $b$th bin with a positive weight $\psi_{k,b}(x_k)$. Note that each sample generally votes for multiple bins with different non-negative weights. The histogram of the class $k$ can then be calculated by normalizing all the votes with the number of samples. Such a process repeats for all classes to create the $K\times B$-dimensional histogram features for a likelihood map or vector. Figs. \ref{fig:histogram} (b) and (c) show an example of the histogram basis functions for class $k$ with $B=6$ bins.  The sample $A$ with the class score $x_A$ would vote for two neighboring bins with non-negative weights $\psi_{k,3}(x_A)$ and $\psi_{k,3}(x_A)$ respectively. 

Unlike the indication functions of the conventional histogram, the linear basis functions for our histogram layer are piecewise differentiable. They can BP errors to lower neural layers, and can calculate the gradients of bin centers and bin widths according to the errors $E$ from its following layers,
\begin{align}
\dfrac{\partial E}{\partial w_{k,b}}=
\begin{cases}
\vert x_k-\mu_{k,b} \vert, & \psi_{k,b}(x_k)>0 , \\
0, & \textnormal{otherwise.} 
\end{cases}
\end{align}
\begin{align}
\dfrac{\partial E}{\partial u_{k,b}}=
\begin{cases}
w_{k,b}, &  \psi_{k,b}(x_k)>0 ~ \textnormal{and}~ x_k-\mu_{k,b}>0,  \\
-w_{k,b}, &  \psi_{k,b}(x_k)>0 ~ \textnormal{and}~ x_k-\mu_{k,b}<0,  \\
0, & \textnormal{otherwise.} 
\end{cases}
\end{align}
$\mu_{k,b}$ and $w_{k,b}$ could then be updated by Stochastic Gradient Descent (SGD). Note that the bin centers and bin widths for different classes are independently learned to capture different data statistics of the classes.

\subsubsection{Learnable histogram layer as existing CNN layers.}
\label{subsubsct:adp} 
\begin{figure}[t]
\centering
\begin{tabular}{c@{\hspace{-2mm}}c@{\hspace{1mm}}c@{\hspace{1mm}}c@{\hspace{1mm}}c}
      \includegraphics[height=1.6cm]{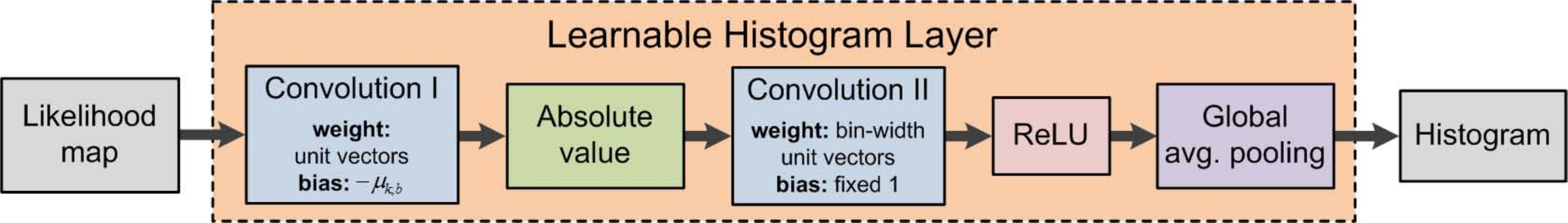}
\end{tabular}
   \caption{Modeling the learnable histogram layer as a stack of existing CNN layers.}\label{fig:histogram-layer}
\end{figure}

One nice property of our proposed learnable histogram layer is that it can be modeled by stacking existing CNN layers, which significantly lowers the implementation difficulty. Such implementation is illustrated in Fig. \ref{fig:histogram-layer}. The input of the histogram layer is a likelihood map or vector from the classification layer, and the output is a $K\times B$-dimensional histogram feature vector. In this subsection, an input likelihood vector is treated as a likelihood map with one spatial dimension equalling $1$.

The operation $x_k-\mu_{k,b}$ for class $k$ and a bin centered at $\mu_{k,b}$ is equivalent to convolving the input likelihood map by a fixed $1\times 1$ kernel $H^\textnormal{I}_{k,b} \in {\bf R}^{K}$ and a learnable bias term $-\mu_{k,b}$ (``Convolution I'' in Fig. \ref{fig:histogram-layer}). Each $1\times 1$ kernel $H^\textnormal{I}_{k,b}$ is a fixed unit vector,

\begin{align}
H^\textnormal{I}_{k,b}(c)=
\begin{cases}
1, & c=k, \\
0, & \textnormal{otherwise,}
\end{cases}
\end{align}
where $H^\textnormal{I}_{k,b}(c)$ denotes the $c$th entry of the kernel $H^\textnormal{I}_{k,b}$.
For class $k$, by convolving the likelihood maps with $B$ such filters we can obtain $B$ new score maps $x_k-\mu_{k,b}$. Since we build a histogram for all $B$ bins, the similar convolutions with different unit-vector kernels and learnable biases would be applied to the input likelihood map for $B$ times. We obtain $K\times B$ new score maps in total, each of which records the results for one bin of a class, and their spatial sizes remain the same. After taking the absolute value of the new score maps, we apply another set of convolutions (``Convolution II'' in Fig. \ref{fig:histogram-layer}) with a total of $K\times B$ learnable $1\times 1$ kernels $H^\textnormal{II}_{k,b} \in {\bf R}^{K \cdot B}$ and fixed bias terms $1$ modeling the operation of $1-|x_k-\mu_{k,b}|\times w_{k,b}$. $|x_k-\mu_{k,b}|$ is the output feature map by the absolute value layer. Each convolution kernel $H^\textnormal{II}_{k,b} $ is a scaled unit vector that models the learnable bin width $w_{k,b}$ for the $b$th bin of class $k$ of the histogram as

\begin{align}
H^\textnormal{II}_{k,b}(c)=
\begin{cases}
w_{k,b}, & c = (k-1)B+k, \\
0, & \textnormal{otherwise.} 
\end{cases}
\end{align}

The $\max\{0, \cdot\}$ function in Equation (\ref{fcn:hist}) is equivalent to the Rectifier Linear Unit (ReLU) non-linearity layer, which takes the feature maps by the ``Convolution II'' as input. The final learnable histogram feature is then obtained by conducting a channel-wise global average pooling on the resulting feature maps.

When training the learnable histogram layer, we ``lock'' the filters for $H^\textnormal{I}_{k,b}$ and $H^\textnormal{II}_{k,b}$ so that only the non-zero entries of them are updated. In this way, we keep the physical meaning of the histogram. These non-zero entries of the filters and the bias terms are not shared across channels, which makes learning bin centers and bin widths for each category independent. We tested abandoning the physical meaning of histograms and allowing the network to freely update all parameters of both convolution filters, which results in inferior performance than our ``locked'' filters (see investigations in Section \ref{subscn:ablation}).

\subsection{Concatenating the histogram features}

Features from our learnable histogram layer capture the global semantic context of the stage-1 likelihood maps or vectors. However, it might not be linearly separable compared with the features by the previous topmost convolution layer. Therefore, we feed the histogram feature into another fully connected layer.  In this paper, we fix the output channels of this layer to be $K\times B$. The output feature is then concatenated to the previous topmost features of all the samples in the same image (i.e., pixel-wise concatenation for semantic segmentation or box-wise concatenation for object detection.) for predicting stage-2 likelihood map or vector (see Fig.\ref{fig:framework} for illustration). 

\subsection{Training schemes}\label{subscn:initialhist}
Our two HistNet models are finetuned based on pre-trained base models (i.e., the VGG-FCN for semantic segmentation and faster-RCNN for object detection) in an incremental manner with 2 phases. In the first phase, only the newly added FC layers are finetuned, with the base models and the learnable histogram layer fixed. The bin centers and widths for each class are initially set as $w_{k,b} = 0.2$ and $\mu_k=\{0,\,0.2,\,0.4,\,0.6,\,0.8,\,1\}$.  In the second phase, we jointly finetune all the layers, with the exception of the above mentioned convolution layers in the learnable histogram layer, which update only their non-zero entries.

\section{Experiments on semantic segmentation}
\subsection{Experimental setup}

We evaluated the proposed HistNet-SS on the semantic segmentation task. The HistNet-SS adopted the VGG-FCN model in \cite{long2014fcn} as the base model for generating stage-1 likelihood maps. 
The base model is initialized by the weights from a VGG-19 model pretrained on ImageNet \cite{ILSVRC15} classification dataset.  Following  \cite{long2014fcn}, we first train the coarse FCN-32s version and use its weights to initialize the final FCN-16s version. All the upsampling deconvolution layers were initialized as bilinear interpolation and allowed adaptation during training. All the new convolutional layers for classification were initialized by Gaussians with zero mean and a standard deviation of $0.01$ and constant biases of $0$.

During training, we adopted the mini-batch Stochastic Gradient Descent (SGD) to optimize
the CNN models and used a mini-batch of 10 images for the semantic segmentation task and 2 images for the object detection task, respectively. We used a gradually decreasing learning rate starting from $10^{-2}$ with a stepsize of 20,000 and a momentum of 0.9.

\subsection{Datasets and evaluation metrics}
We evaluate the proposed HistNet-SS on the SIFTFlow \cite{liu2008sift}, Stanford background \cite{gould2009decomposing} and PASCAL VOC 2012 \cite{pascal_voc_2012} segmentation datasets. 
The SIFTFLow dataset consists of 2488 training images and
200 test images. All the images are of size $256\times 256$ and contain 33 semantic labels. 
The Stanford background dataset contains 715 images of outdoor scenes composed of 8 classes. Following the train/test split in \cite{socher2011parsing,sharma2015deep},  572 images are selected as the training set and the rest 143 images as the test set.
PASCAL VOC datasets consists of 1464, 1449, and 1456 images for training, validation, and testing, respectively. The dataset is augmented by the extra annotations provided by \cite{hariharan2011semantic}, resulting in 10582 training images. 
For the first two dataset, we augmented the training set by randomly scaling, rotating and horizontally flipping each training image for 5 times. The scaling factors and the rotation angles were randomly chosen in the ranges of [0.9, 1.1] and [$-8^\circ, 8^\circ$]. For PASCAL VOC dataset, we did not conduct data augmentation. No class balancing is performed on any dataset. 

Following the common practice, we evaluate the compared methods by the per-pixel and per-class accuracies on SIFTFlow and Stanford background datasets. For PASCAL VOC 2012 segmentation dataset, The performance is measured in terms of intersection-over-union (IOU) averaged across the 21 classes.

\subsection{Overall performance}
\subsubsection{SIFTFlow dataset}
\begin{figure}[t]
\centering
\begin{tabular}{c@{\hspace{2mm}}c@{\hspace{1mm}}c@{\hspace{1mm}}c@{\hspace{1mm}}c@{\hspace{1mm}}c}
   \includegraphics[height=2.5cm]{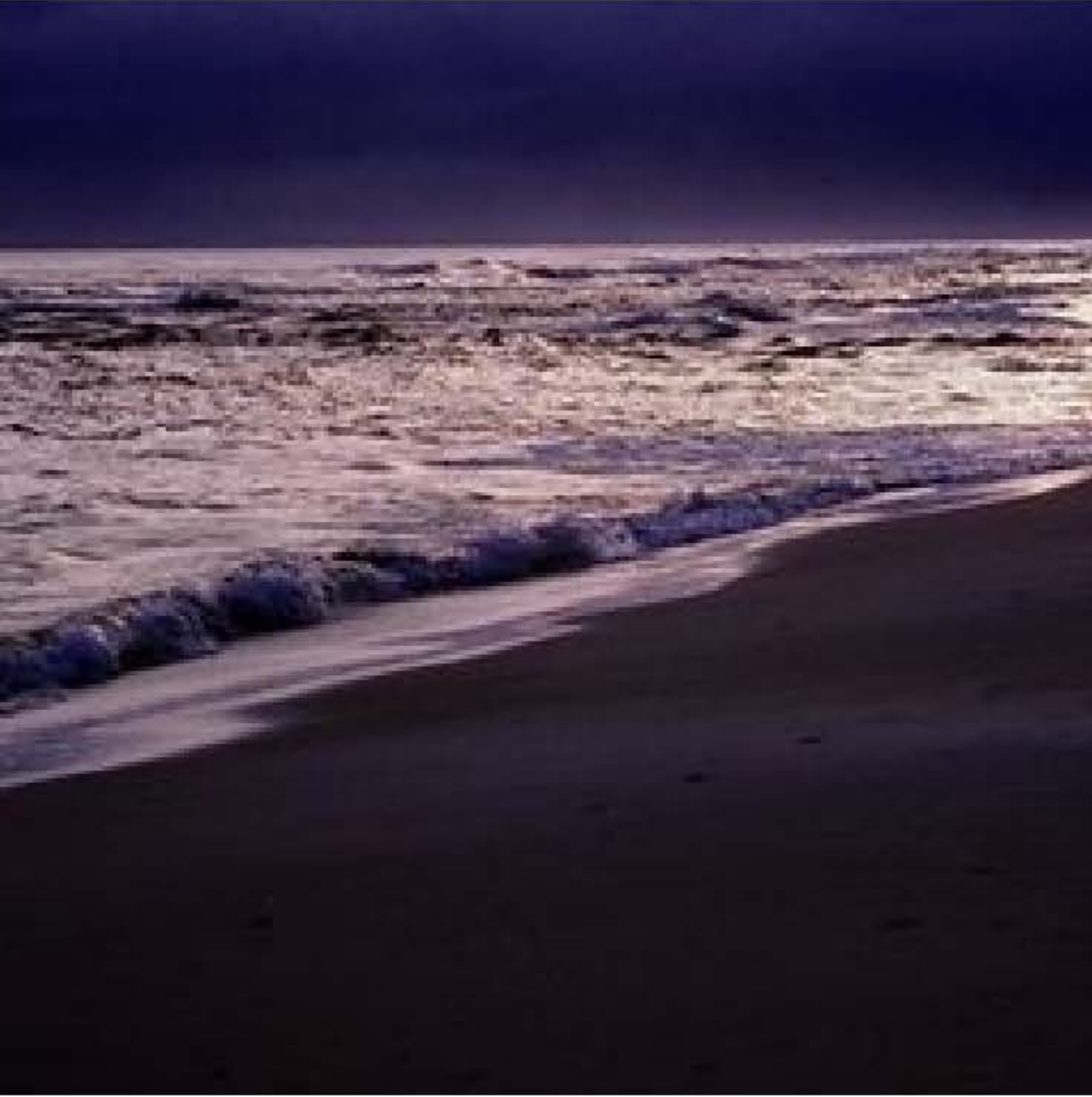}&    \includegraphics[height=2.5cm]{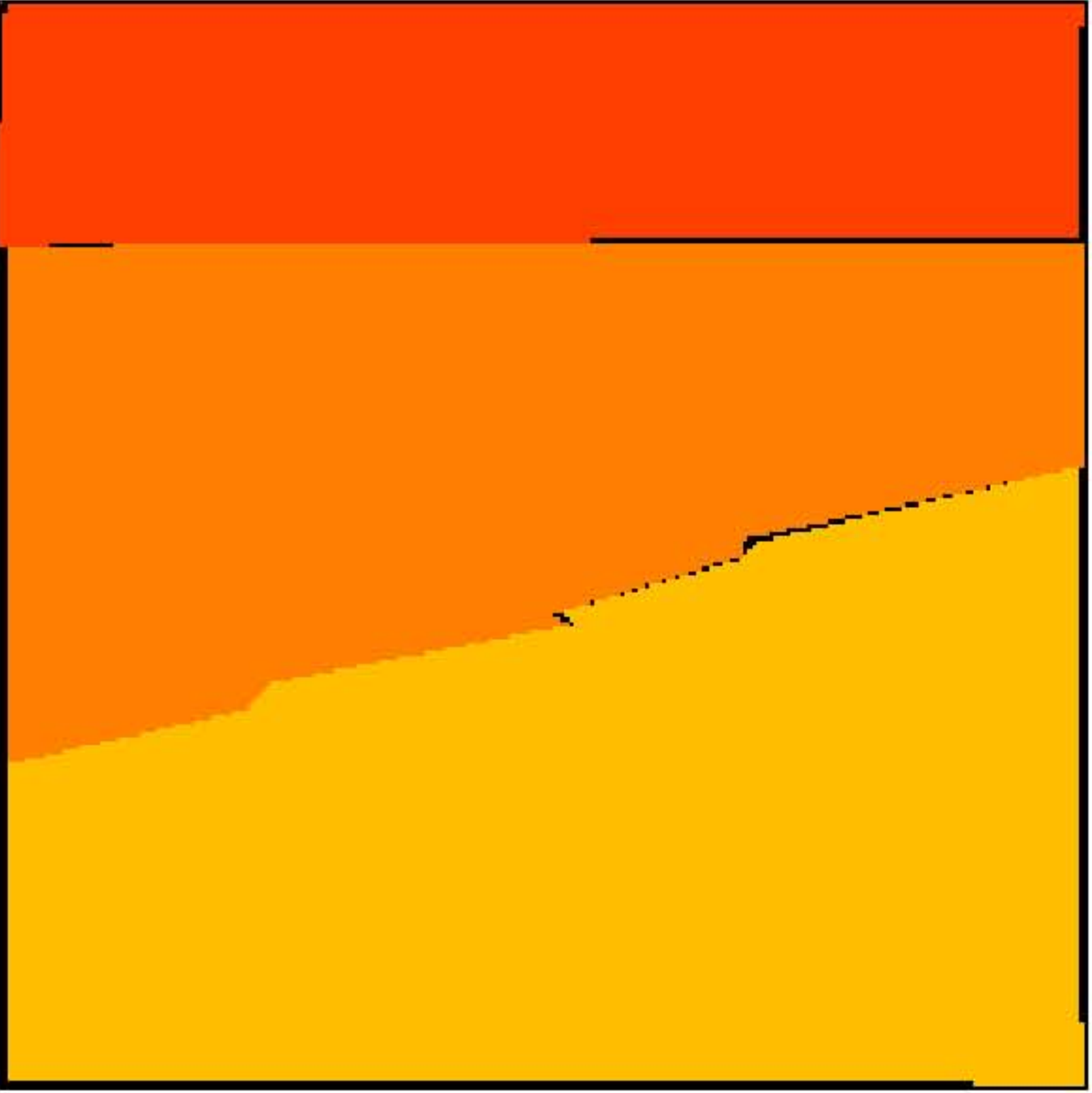} &
\includegraphics[height=2.5cm]{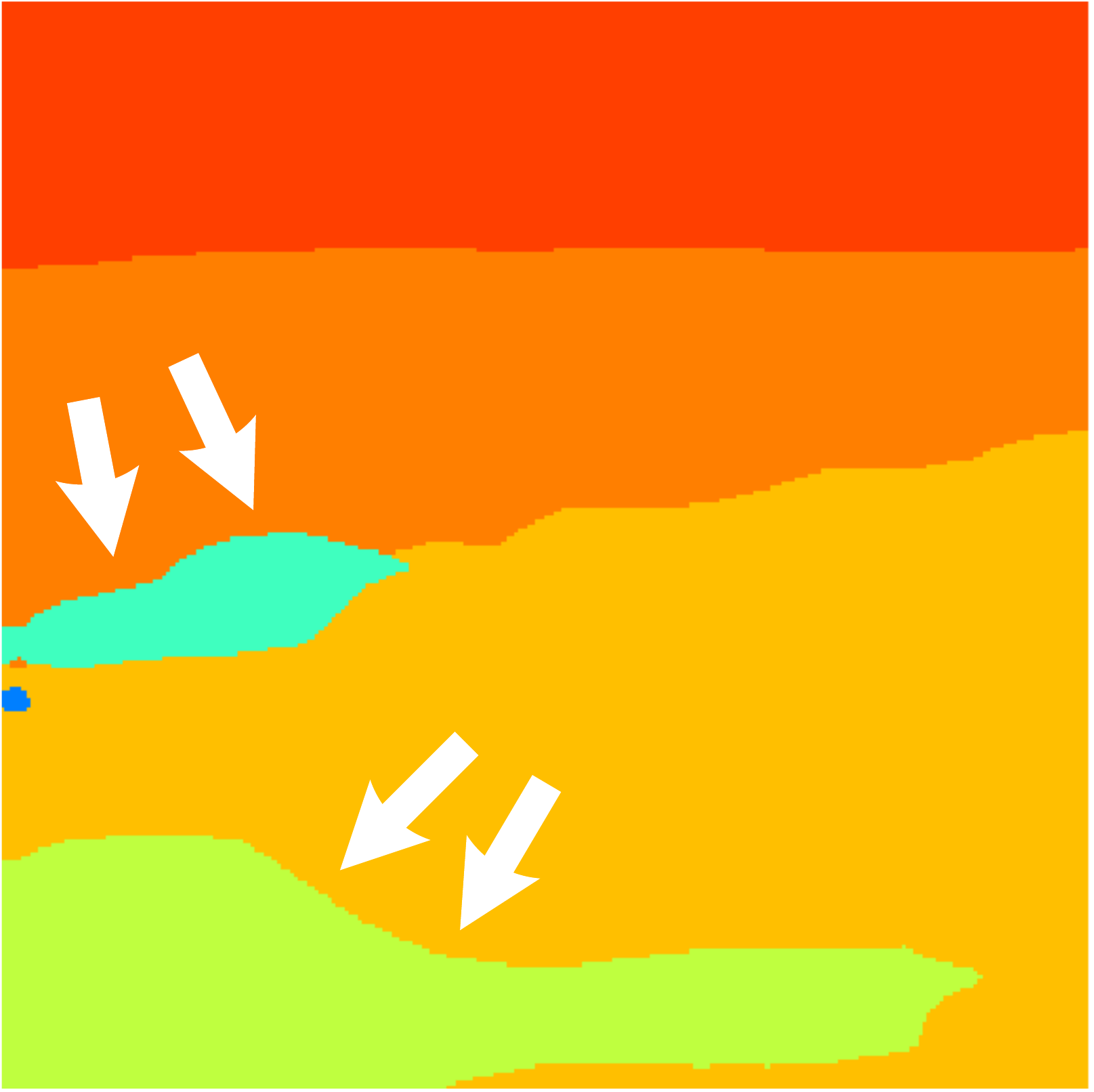} &
\includegraphics[height=2.5cm]{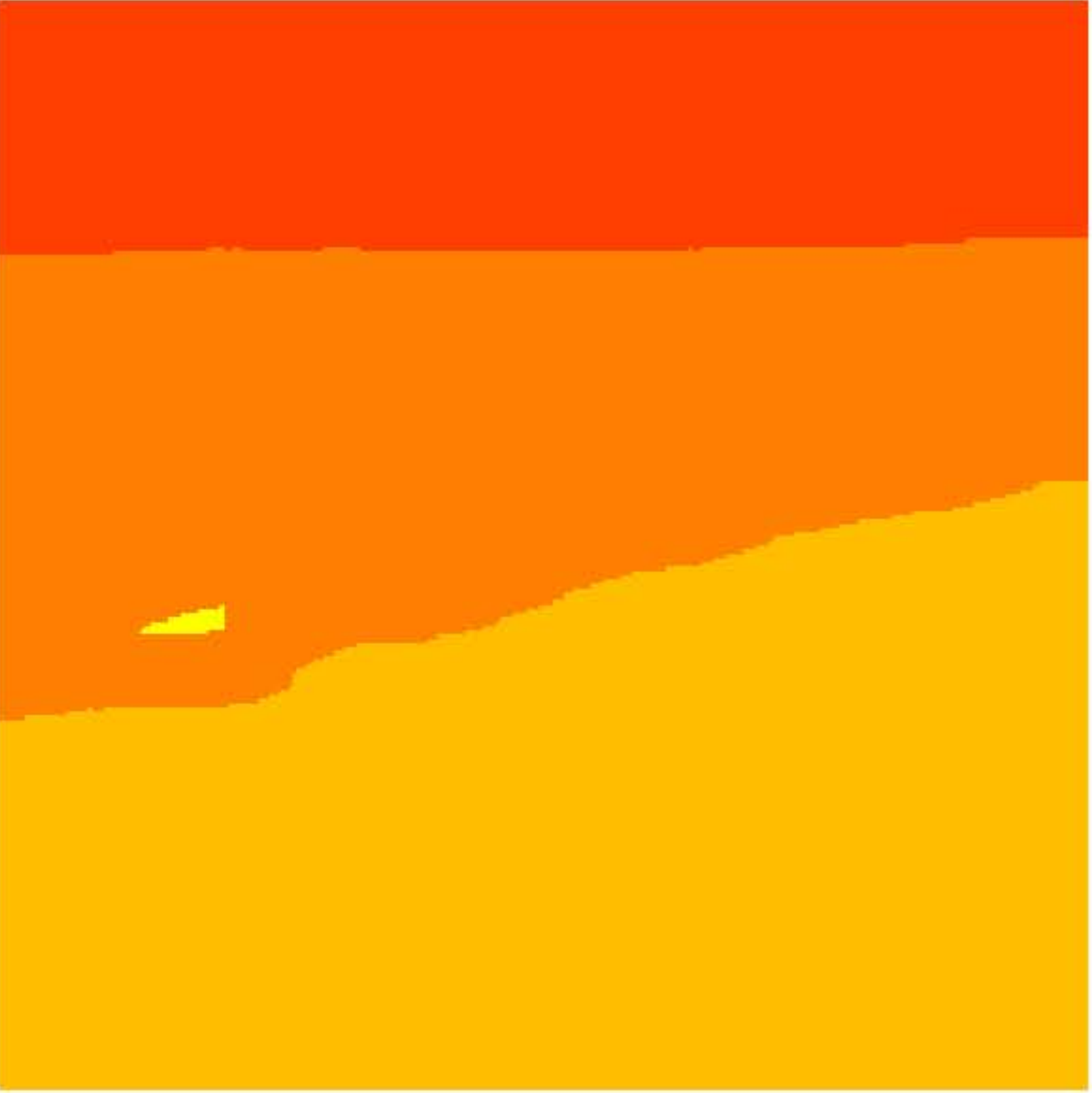} &
\includegraphics[height=2.5cm]{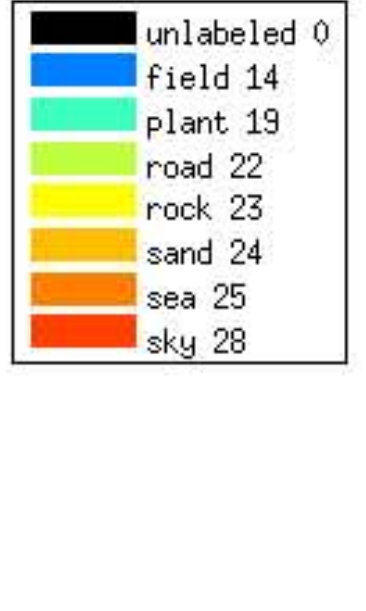} &
   \\
   \includegraphics[width=2.5cm]{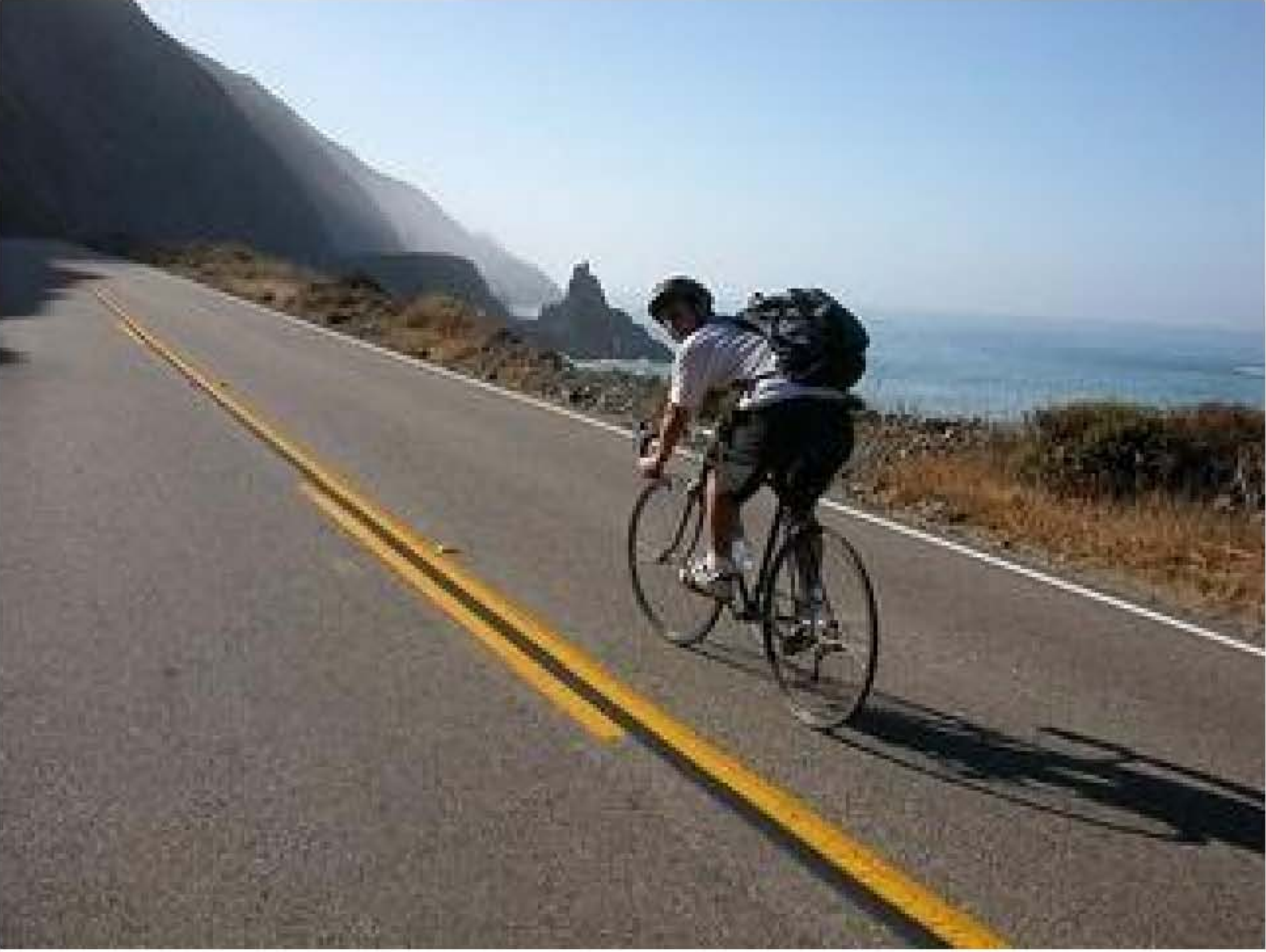}&    \includegraphics[width=2.5cm]{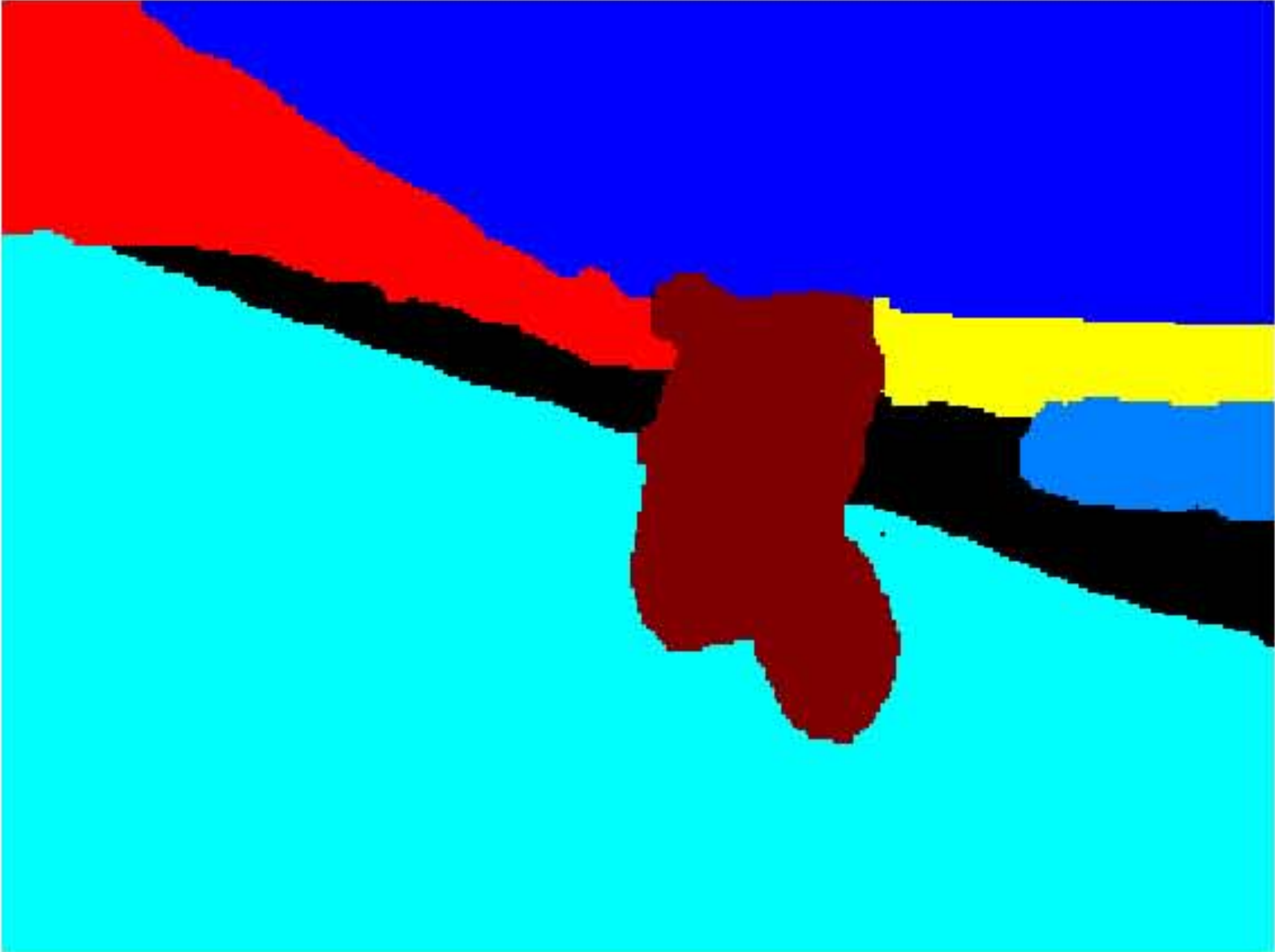} &
\includegraphics[width=2.5cm]{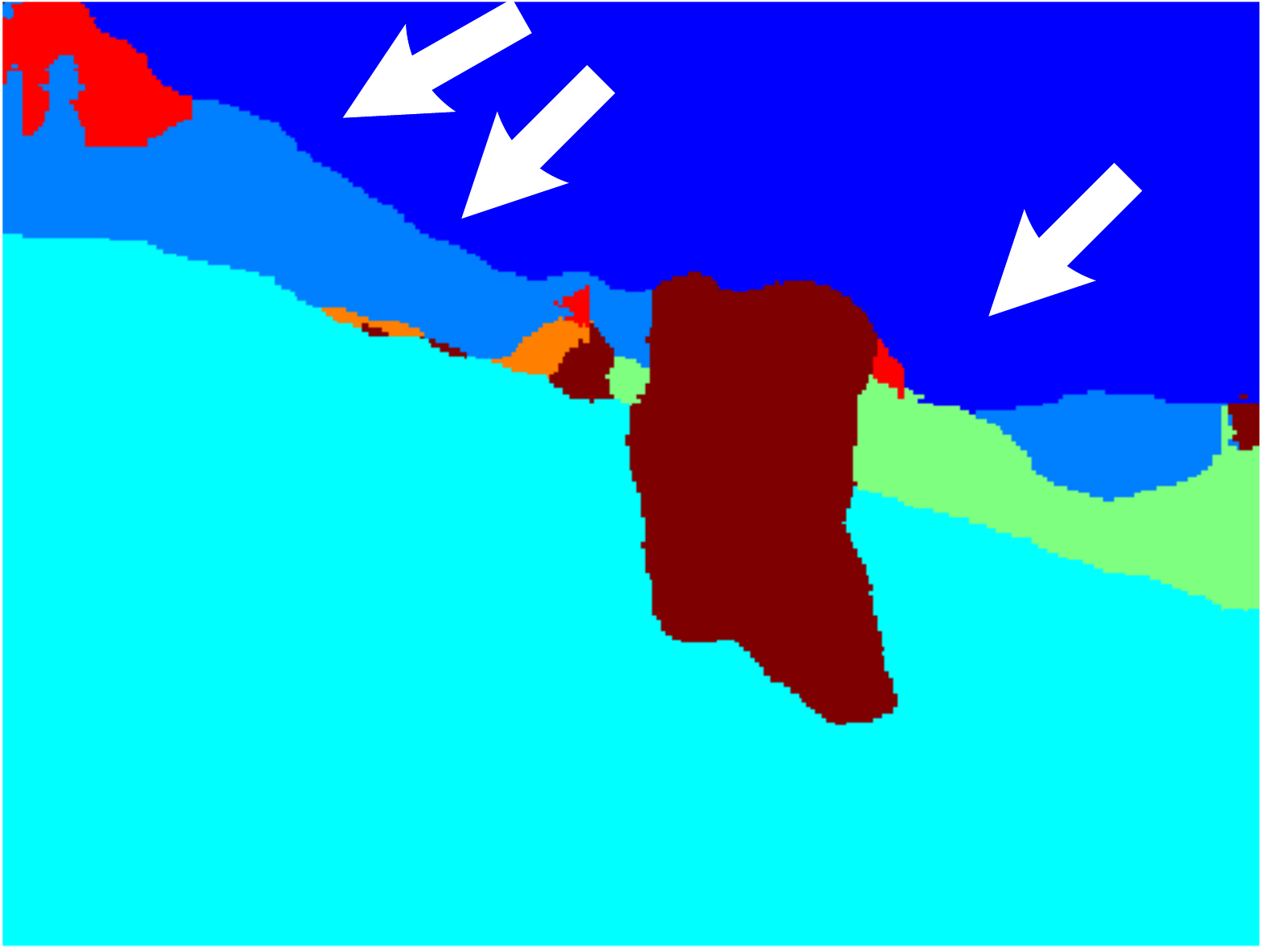} &
\includegraphics[width=2.5cm]{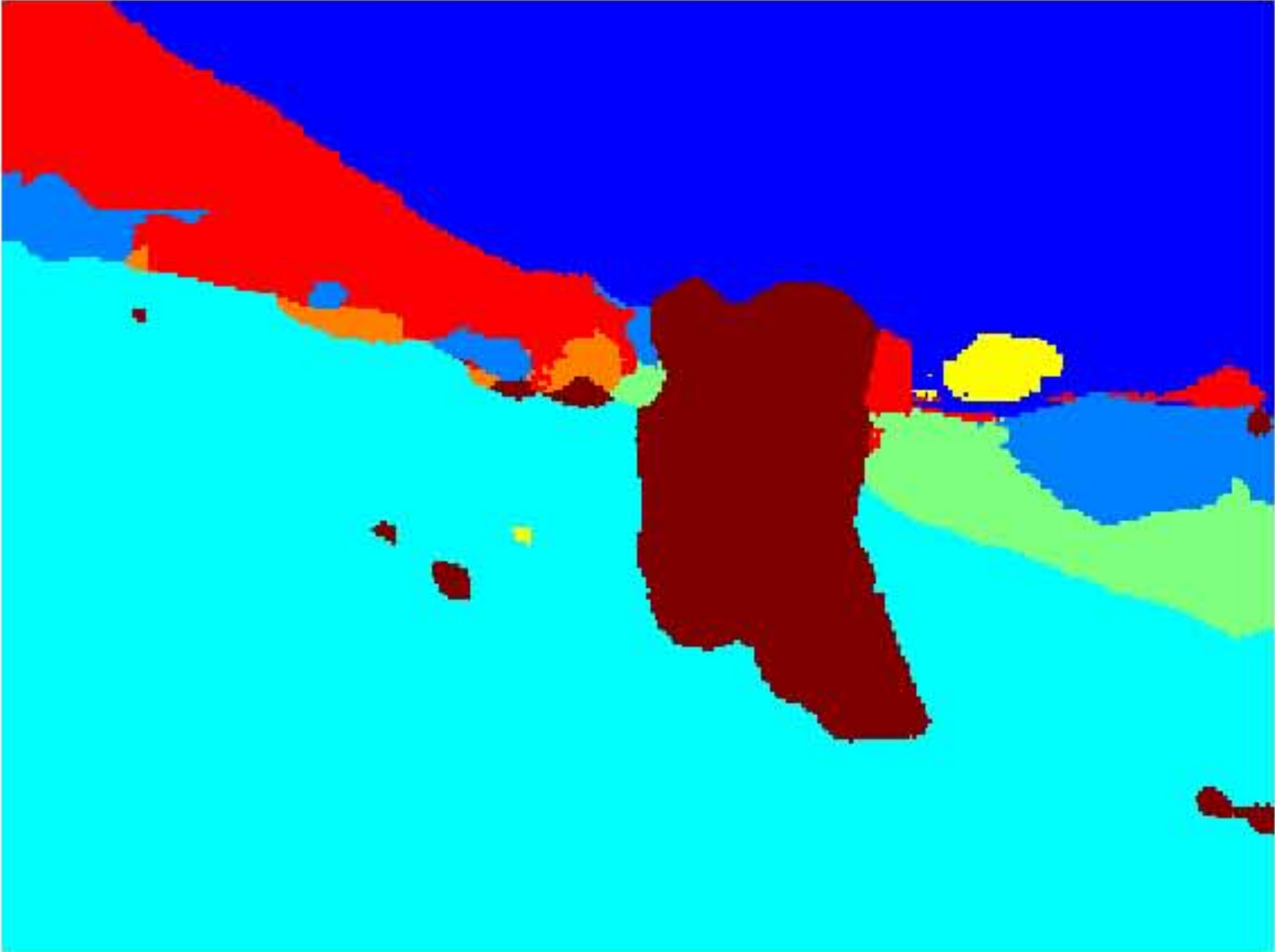} &
\includegraphics[width=1.5cm]{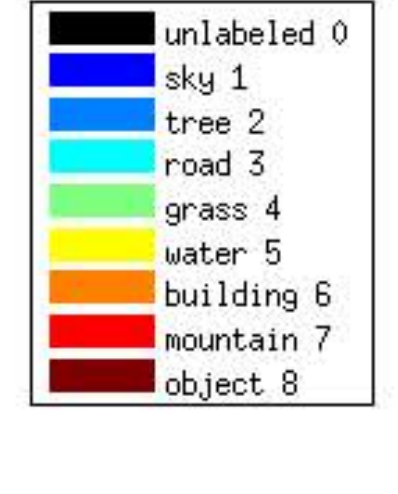} &
   \\
   Image & Label & FCN & HistNet-SS (ours)
\end{tabular}
   \caption{Example results on scene labeling by our HistNet-SS. 
   (Row 1) HistNet-SS successfully predicts most erroneous ``plant'' and ``road'' pixels by FCN to ``sea'' and ``sand'', which are more likely to appear at sea shore.  (Row 2) HistNet-SS labels most ``mountain'' pixels correctly and discovers some ``water'' pixels not found by FCN. 
   }\label{fig:results}
\end{figure}

For the SIFTFlow dataset, we compared our method with state-of-the-art methods, which include both
deep-learning-based \cite{farabet2013learning,pinheiro2013recurrent,sharma2015deep,eigen2015predicting,long2014fcn},   and non-deep-learning-based methods \cite{tighe2010superparsing,liu2008sift,yang2014context}. The accuracies by different methods are reported in Table \ref{tbl:siftflow_result}(a). The HistNet-SS achieves state-of-the-art performance. Note that here the
HistNet-SS is based on the FCN model implemented by ourselves. This FCN baseline achieves a higher per-pixel (0.86 v.s. 0.851) accuracy but a lower per-class accuracy  (0.457 v.s. 0.517) compared to the results reported in \cite{long2014fcn}, which might result from different data distribution caused by our data augmentation. The HistNet-SS is initialized by our implemented FCN model. Some qualitative results are shown in Fig. \ref{fig:results}.

As shown in Table \ref{tbl:siftflow_result}(a), with the learnable histogram layer, HistNet outperformed its VGG-FCN base model by $1.9\%$ and $5\%$ for per-pixel and per-class accuracies, respectively. 
The base model of HistNet (denoted as HistNet-SS stage-1) has exactly the same net structure with the VGG-FCN but is jointly finetuned within the HistNet-SS. It is interesting to see that the prediction by the base model also benefited from the joint finetuning.  This demonstrates that the bottom convolution layers now learn better feature representations, while keeping the same model complexity and without extra training data or supervision. 

\begin{table}[t!]
\scriptsize
\caption{Per-pixel and per-class accuracies on (a) the SIFTFlow dataset and (b) the Stanford background dataset by different methods. Best accuracies are marked in bold.}\label{tbl:siftflow_result}
\begin{minipage}[b]{0.5\linewidth} 
\centering
\begin{tabular}{|l|c|c|}
\hline
Methods & Per-pixel & Per-class\\ \hline \hline
Tighe et al. \cite{tighe2010superparsing}    & 0.769     & 0.294     \\
Liu et al. \cite{liu2008sift}         & 0.748     & n/a        \\
Farabet et al. \cite{farabet2013learning}                      & 0.785     & 0.296     \\ 
Pinheiho et al. \cite{pinheiro2013recurrent}          & 0.777     & 0.298     \\ 
Sharma et al. \cite{sharma2015deep}         & 0.796     & 0.336     \\ 
Yang  et al. \cite{yang2014context}                  & 0.798     & 0.487     \\
Eigen et al. \cite{eigen2015predicting} & 0.868 & 0.464 \\
\hline \hline
FCN \cite{long2014fcn}                  & 0.851     & \textbf{0.517} \\
FCN (our implement)  & 0.860  & 0.457\\
FCN+FC-CRF  & 0.865  & 0.468\\
HistNet-SS stage-1 & 0.876  & 0.505\\
HistNet-SS  & \textbf{0.879} & 0.5 \\ 
HistNet-SS+FC-CRF & \textbf{0.879} &  0.512\\\hline
\end{tabular} \\
(a) SIFTFlow dataset
\end{minipage}%
 \begin{minipage}[b]{0.5\linewidth}
 \centering
\begin{tabular}{|l|c|c|}
\hline
Method & Per-pixel & Per-class \\ \hline \hline
Gould et al. \cite{gould2009decomposing}     & 0.764     & n/a       \\
Tighe et al. \cite{tighe2010superparsing}    & 0.775     & n/a     \\
Socher et al. \cite{socher2011parsing}    & 0.781     & n/a     \\
Lempitzky et al. \cite{lempitsky2011pylon}  & 0.819     & 0.724     \\
Farabet  et al. \cite{farabet2013learning}                      & 0.814     & 0.76    \\ 
Pinheiho et al. \cite{pinheiro2013recurrent}          & 0.802     & 0.699     \\ 
Sharma et al. \cite{sharma2015deep}         & 0.823     & 0.791     \\ \hline \hline
FCN (our implement) & 0.851 & 0.811\\
FCN+FC-CRF & 0.862 & 0.82\\
FCN+MOPCNN \cite{gong2014multi} & 0.863 & 0.811 \\
HistNet-SS stage-1 & 0.871 & \textbf{0.838} \\ 
HistNet-SS & 0.871 & 0.837 \\
HistNet-SS+FC-CRF &  {\bf 0.881} & 0.837  \\ \hline	
\end{tabular}\\
(b) Stanford background dataset
\end{minipage}%
\end{table}

\subsubsection{Stanford background dataset}
The results on the Stanford background dataset are reported in Table \ref{tbl:siftflow_result}.(b). 
Since FCN \cite{long2014fcn} did not report their results on this dataset, here we only report the results of FCN implemented by ourselves, which surpasses state-of-the-art methods.  Our proposed HistNet-SS achieves the best performance with both evaluation metrics, which shows the effectiveness of incorporating the global histogram layer into the network. 
We also evaluated the performance of the HistNet-SS stage-1, i.e., the base model after jointly finetuning with the proposed learnable histogram layer. Its performance is slightly better than the final combined result.  This may be an evidence that the HistNet-SS does not simply improve its performance by adding more parameters to fit the dataset. 
On the contrary, it helps bottom convolution layers learn more discriminative features with statistical context features. 

We also compared HistNet-SS with Gong et al. \cite{gong2014multi}, which also used global features in a CNN framework. We used their code to extract image-level features by an ImageNet-pretrained AlexNet model \cite{krizhevsky2012imagenet}. Then the off-the-shelf feature is repeatedly concatenated to the original feature maps at each location, followed by a newly trained classifier. As shown in Table \ref{tbl:siftflow_result}, the result FCN + MOPCNN is inferior than HistNet-SS, since it cannot be trained in an end-to-end manner.  

In addition, we also tried to utilize the fully-connected CRF algorithm \cite{koltun2011efficient} to regularize the output likelihood map by our HistNet-SS following \cite{chen2014semantic}. The accuracies on both SIFTFlow and Stanford background datasets could be further improved, which demonstrate that our histogram context features are complementary to the semantic context modeled by graphical models.

\subsubsection{PASCAL VOC 2012 segmentation dataset}
We also trained HistNet-SS based on the publicly-available DeepLab model (multi-scale features and large field-of-view) \cite{chen2014semantic} with the augmented ``train'' set. DeepLab \cite{chen2014semantic} achieves a $64.2\%$ mean IOU, while our method HistNet-SS improves it to $67.5\%$. It shows that the HistNet-SS benefits from the learned histogram of foreground objects categories.

\subsection{Investigation on the HistNet-SS}\label{subscn:ablation}
To further verify the effectiveness of the HistNet-SS, we designed multiple baseline networks to analyse each component of our learnable histogram layer.
 
\begin{table}
\scriptsize
\centering\caption{Performance of different baseline models of the HistNet-SS and their corresponding numbers of extra parameters.}
\begin{tabular}{|l|c|c||c|c|c|}
\hline
 \multirow{2}{*}{Methods} & \multicolumn{2}{c||}{SIFTFlow} & \multicolumn{2}{c|}{Stanford background} & \multirow{1}{*}{$\#$ extra parameters} \\ \cline{2-5}
 & per-pixel & per-class & per-pixel & per-class & (SIFTFlow/Stanford) \\ \hline
FCN baseline & 0.860 & 0.450 & 0.851 & 0.811  & 0\\ \hline
FCN-fix-hist & 0.872 & 0.481  & 0.860  & 0.829 &  $\sim190,000$ /  $36,000$ \\ \hline
FCN-free-all & 0.870  & 0.489 & 0.862 & 0.824  & $\sim190,000$ /  $36,000$\\ \hline
FCN-fc7-global & 0.870 & 0.462 & -& -& $\sim$ 960,000 / 23,000 \\ \hline
FCN-score-global & 0.873  & 0.480 & 0.863 & 0.825 & $\sim$ 150,000 / 35,000 \\ \hline
R-HistNet-SS & {\bf 0.880} & 0.486 & {\bf 0.872} & {\bf 0.845}& $\sim$ $380,000 $ /  $72,000$  \\ \hline\hline
HistNet-SS (ours) & 0.879 &  {\bf 0.5} &  0.871 & 0.837  & $\sim$ $190,000$ /  $36,000$ \\ \hline
\end{tabular}
\label{tbl:ablation-sl}
\end{table}

\subsubsection{Learnable histogram v.s. fix-bin histogram v.s. ``unlocked histogram''.} 
In order to find out whether we can benefit from learning histogram bin centers and bin widths, and whether keeping the physical meaning of the histogram helps training, we designed two baselines, FCN-fix-hist and FCN-free-all. They were both initialized  in the same way as HistNet-SS. For FCN-fix-hist, we fixed its bin centers and widths during training. Recall that for HistNet-SS, we ``locked'' the $1\times 1$ kernels to make it only update the non-zero entries. For FCN-free-all, we ``unlocked'' all the convolution kernels and biases in the learnable histogram layer so that they could adapt freely.  It no longer holds the physical meaning of a histogram. As shown in Table \ref{tbl:ablation-sl}, FCN-fix-hist is not as good as our HistNet-SS, which confirms our assumption that a learnable histogram is critical to better describe the context. FCN-free-all performs inferiorly to HistNet-SS by a small margin. It may suggest that keeping the physical meaning of the histogram acts as a regularizer which has fewer learnable parameters to avoid overfitting.

\subsubsection{Statistical context v.s. non-statistical context.}
To verify whether statistical context is better than non-statistical context, we trained two different baseline networks. FCN-fc7-global feeds the VGG-FCN fc7 layer's output, i.e., the topmost feature maps, to a $1\times 1$ convolution layer with $K\times B$ output feature maps to match the HistNet-SS. It applies global average pooling first and then concatenates the same vector at each location of the topmost feature maps, followed by a fully connected classification layer. FCN-score-global is similar to FCN-fc7-global, except it takes the likelihood maps as input. The numbers of extra parameters are recorded in Table \ref{tbl:ablation-sl}. The HistNet-SS has the fewest extra parameters among the settings. In Table \ref{tbl:ablation-sl} it can be seen that FCN-score-global and FCN-fc7-global perform comparably. However, they are inferior to the HistNet-SS. 
We also tried adding another learnable histogram layer to the stage-2 likelihood maps to form a recurrent HistNet (denoted as R-HistNet-SS), which is initialized by HistNet-SS and its prediction is based on the average of three likelihood maps. However, no significant improvement is observed. 
\begin{table}[t]\caption{Results of object detection (mAP \%) on the VOC 2007 test dataset. RCNN and fast RCNN results are from \cite{girshick2015fast}.}
\centering
\scriptsize
\label{my-label}
\begin{tabular}{cccccccccccc}
\hline
\multicolumn{1}{|l|}{Methods}            & aero      & bike & bird  & boat     & bottle & bus   & car   & cat  & chair & cow  & \multicolumn{1}{l|}{}     \\ \hline
\multicolumn{1}{|l|}{RCNN  \cite{rirshick2014rich}}        & 73.4      & 77.0 & 63.4  & 45.4     & 44.6   & 75.1  & 78.1  & 79.8 & 40.5  & 73.7 & \multicolumn{1}{l|}{}     \\ 
\multicolumn{1}{|l|}{fast RCNN \cite{girshick2015fast}}   & 74.5      & 78.3 & 69.2  & 53.2     & 36.6   & 77.3  & 78.2  & 82.0 & 40.7  & 72.7 & \multicolumn{1}{l|}{}     \\
\multicolumn{1}{|l|}{faster RCNN \cite{renNIPS15fasterrcnn}} & 69.1     & 78.3 & 68.9  & 55.7     & 49.8   & 77.6  & 79.7  & 85.0 & 51.0  & 76.1 & \multicolumn{1}{l|}{}     \\ 
\multicolumn{1}{|l|}{HistNet-OD stage-1}  &   68   &  80.3     &  74.1     & 55.7         & 53.3 & 83.6      &  80.2     & 85.1 & 53.7       & 74.2     & \multicolumn{1}{l|}{}     \\ 
\multicolumn{1}{|l|}{HistNet-OD}  &   67.6    &  80.3     &  74.1     & 55.6         & 53.2 & 83.4     &  80.2     & 85.1 & 53.6       & 74     & \multicolumn{1}{l|}{}     \\ \hline \hline
\multicolumn{1}{|l|}{}            & table     & dog  & horse & mbike & person & plant & sheep & sofa & train & tv   & \multicolumn{1}{l|}{\textbf{mAP}}  \\ \hline
\multicolumn{1}{|l|}{RCNN \cite{rirshick2014rich} }        & 62.2      & 79.4 & 78.1  & 73.1     & 64.2   & 35.6  & 66.8  & 67.2 & 70.4  & 71.1 & \multicolumn{1}{l|}{66.0} \\ 
\multicolumn{1}{|l|}{fast RCNN \cite{girshick2015fast}}   & 67.9      & 79.6 & 79.2  & 73.0     & 69.0   & 30.1  & 65.4  & 70.2 & 75.8  & 65.8 & \multicolumn{1}{l|}{66.9} \\
\multicolumn{1}{|l|}{faster RCNN \cite{renNIPS15fasterrcnn}} & 64.2      & 82.0 & 80.5  & 76.2     & 75.8   & 38.5  & 71.4  & 65.4 & 77.8  & 66.1 & \multicolumn{1}{l|}{69.5} \\ 
\multicolumn{1}{|l|}{HistNet-OD stage-1} & 69.3 & 82.5      & 84.9       & 76.5     &       77.7 & 44.2      &   71.7    & 66.6    & 75.5      &    71.8 & \multicolumn{1}{l|}{71.4}       \\ 
\multicolumn{1}{|l|}{HistNet-OD}   & 69.3 & 82.5      &  84.8       & 76.3     &   77.6 & 44.1      &   71.9    & 66.8     & 75.4      &    71.9 & \multicolumn{1}{l|}{71.4}   \\ \hline
                                 &           &      &       &          &        &       &       &      &       &      &                          
\end{tabular}
\label{table:voc07_det}
\end{table}

\section{Experiments on object detection}
\subsection{Experimental setting}
We adopted the faster-RCNN \cite{renNIPS15fasterrcnn} pipeline to build the proposed HistNet-OD model and evaluated it on the PASCAL VOC 2007 detection benchmark \cite{everingham2007pascal}. 
This dataset consists of about 5k trainval images and 5k test images over 20 categories. The standard evaluation metric is the mean Average Precision (mAP).
We utilized the faster-RCNN model trained by its python interface, which is provided by the authors of \cite{renNIPS15fasterrcnn}. It has a slightly lower mAP than the MATLAB version one reported in their paper (0.695 v.s. 0.699).
HistNet-OD stage-1 is initialized by this model. The histogram layer parameters are initialized as mentioned in Section \ref{subsubsct:adp}.  The new fully connected layers were initialized by zero-mean Gaussian with a standard deviation of 0.01.  We finetuned the HistNet-OD with the VOC07 trainval set and tested it with the VOC07 test set.

\subsection{Overall performance}
We report the overall performance of the HistNet-OD on the VOC 2007 test dataset. As shown in Table \ref{table:voc07_det}, the HistNet-OD outperforms the faster-RCNN by $1.9\%$.  This result shows that the learnable histogram layer has good generalization ability and can also be applied to the object detection task. Similar to the semantic segmentation task, our base model HistNet-OD stage-1  was also improved by jointly finetuning with the learnable histogram layer, which indicates that the feature representations learned by the base model are also improved.

\subsection{Investigation on the HistNet-OD}
Similar to the experiments in semantic segmentation (Section \ref{subscn:ablation}), we also designed a baseline models, faster-RCNN-fc7-global, to study the influence of statistical context and non-statistical context features. The features of the faster-RCNN's fc7 layer go through a new FC layer, and are concatenated back to the previous topmost features after global average pooling. A new FC layer acting as the classifier is trained on top of the new concatenated features. 

The mAP result of faster-RCNN-fc7-global is 0.704, with 170k extra parameters, compared to 0.714 by HistNet-OD, with only 91k extra parameters. This confirms that the learnable statistical feature outperforms the non-statistical one with fewer parameters. If the histogram parameters of HistNet-OD are fixed, the mAP is 0.707. It shows that HistNet-OD can benefit from tuning the parameters. 



\section{Conclusions}

One interesting observation is that by training with the learnable histogram layer, the base network is also improved by jointly finetuning. Previous works \cite{chatfieldreturn,szegedy2014going,he2015deep} mostly focus on designing deeper networks to have stronger expressive power. However, this work shows that after finetuning with a deeper network, the original base model can also be improved, which may suggest a new way for model training: we can train a deep neural network with learnable histogram layers and multiple loss functions at different layers, and only use the base network for deployment. 

In this work, we proposed a learnable histogram layer for deep neural networks, which does not only back-propagate errors, but also learns optimal bin centers and bin widths. Based on this learnable histogram layer, two models are designed for semantic segmentation and object detection, respectively. Both models show state-of-the-art performance, which demonstrates that the proposed learnable histogram layer is able to learn effective statistical features and is easy to generalize to different domains. In-depth investigations were conducted to analyse the effectiveness of different components of the learnable histogram layer.

\subsubsection{Acknowledgements.}
This work is supported by SenseTime Group Limited, the General Research Fund sponsored by the Research Grants Council of Hong Kong (Project Nos. CUHK14206114, CUHK14205615, CUHK417011,  CUHK419 412, CUHK14203015, and CUHK14207814), the Hong Kong Innovation and Technology Support Programme (No. ITS/221/13FP), National Natural
Science Foundation of China (Nos. 61371192, 61301269), and PhD programs foundation of China (No. 20130185120039).

\clearpage

\bibliographystyle{splncs}
\bibliography{egbib}
\end{document}